\documentclass[pmlr,twocolumn,10pt]{jmlr} 

\usepackage{booktabs}
\usepackage{siunitx}

\newcommand{\equal}[1]{{\hypersetup{linkcolor=black}\thanks{#1}}}

\title[Fairness in Global Health]{Globalizing Fairness Attributes in Machine Learning: A Case Study on Health in Africa}

\author{%
\Name{Mercy Nyamewaa Asiedu}\equal{These authors contributed equally} \Email{masiedu@google.com}\\
\addr Google Research
\AND
\Name{Awa Dieng}\footnotemark[1] \Email{awadieng@google.com}\\
\addr Google Research
\AND
\Name{Abigail Oppong} \Email{abigoppong@gmail.com}\\
\addr Ghana NLP
\AND
\Name{Maria Nagawa} \Email{maria.nagawa@duke.edu}\\
\addr Duke University
\AND
\Name{Sanmi Koyejo} \Email{sanmik@google.com}\\
\addr Stanford University, Google Research
\AND
\Name{Katherine Heller} \Email{kheller@google.com}\\
\addr Google Research
}

\begin{document}

\maketitle

\begin{abstract}
With growing machine learning (ML) applications in healthcare, there have been calls for fairness in ML to understand and mitigate ethical concerns these systems may pose. Fairness has implications for global health in Africa, which already has inequitable power imbalances between the Global North and South.  This paper seeks to explore fairness for global health, with Africa as a case study. We propose fairness attributes for consideration in the African context and delineate where they may come into play in different ML-enabled medical modalities. This work serves as a basis and call for action for furthering research into fairness in global health.
\end{abstract}

\section{Introduction}
\label{sec:intro}

Machine learning (ML) models have the potential for far reaching impact in health. However they also have the potential to propagate biases that reflect real-world historical and current inequities and could lead to unintended, harmful outcomes, particularly for vulnerable populations \citep{huang22, siala22, chenIY21, char18, gianfrancesco18, obermeyer19}. Sources of biases include missing data, non-representative sample size, imbalanced group distributions, and misclassification or measurement error \citep{gianfrancesco18}. 
In health, this can affect vulnerable populations - patients with fractured care, low literacy levels, underrepresented subgroups, and/or patients who access clinical facilities with limited devices for measurements or data input methods \citep{gianfrancesco18, obermeyer19}. A prominent example is an algorithm used in the United States that deprioritized sicker black patients for care, based on a faulty cost variable, which perpetuated structural socio-economic disparities \citep{obermeyer19}. 
This has led to the institution of the algorithmic fairness field, various attempts to correct machine learning biases, through evaluating models and understanding attributes that may cause ML models to make unfair or biased decisions.

There are multiple definitions for global health but we use the definition proposed by \citet{beaglehole10}: \textit{"collaborative trans-national research and action for promoting health for all"}. Fairness is especially important for global health, which despite positive outcomes such as decreased has been plagued with inequitable power imbalances between high-income countries (HICs) and low- and middle-income countries (LMICs) \citep{holst20}. 
The recent decolonizing global health movement sheds light on ``how knowledge generated from HICs define practices and informs thinking to the detriment of knowledge systems in LMICs''\citep{eichbaum21}. 
\citet{eichbaum21} explore intersections between colonialism, medicine and global health, and how colonialism continues to impact global health programs and partnerships. They make the point that even the use of the term ``Global health" to describe health in LMICs is problematic. 
Given that most machine learning models are developed with problem formulation, personnel, resources and data from HICs, and may be imported with little regulation to LMICs, there is a risk for algorithmic colonialism \citep{abeba20, mohamed20}, which would further exacerbate current power imbalances in global health. On the other hand, increasing internet access, and democratizing of machine learning knowledge and tools, present an opportunity to create sustainable change in global health by empowering research partnerships to develop ML tools for locally relevant applications in health. 
As we explore where machine learning in health could be beneficial to global health, we must also proactively identify and mitigate biases they could generate.\\
 
To date, fairness in health research has furthered understanding and mitigation of biases along the machine learning development pipeline \citep{chenIY21, char18, siala22}. However most are contextualized to HICs, there are a few studies that have explored contextualizing fairness beyond the west.

\citet{sambasivan21} utilizes mixed methods studies to identify subgroups, proxies and harms for fair ML in the context of India. For natural language processing applications, \citet{bhatt22} identify India-specific axes for disparities such as region and caste, and redefine global axes in an India context, such as gender, religion, sexual identity and orientation, and ability. \citet{fletcher21} provide a detailed recommendations for fairness, bias and appropriate AI use in global health with examples using pulmonary disease classification in India.\\

This paper builds on previous work by exploring fairness attributes for global health, with Africa as a case study:
\begin{itemize}
    \item We identify fairness attributes that should be considered with respect to Africa, proposing (1) axes of disparities between Africa and the West, and (2) global axes of disparities in the context of Africa. 
    \item We delineate real world ML for health applications where these axes should be considered.
    \item We end with a discussion of contextual challenges and open avenues of research for impactful use of fair ML in Africa.  
\end{itemize}

While there are common themes that may resonate across the continent, Africa is extremely diverse and within there are attributes and axes of disparities that may apply to some countries and not to others. This is early work attempting to lay a foundation, on contextualizing fairness beyond Western contexts using machine learning application to health in Africa as a case study. We hope to convey the importance of this area of work and generate  interest in exploring this in greater detail. 

\section{Fairness Considerations}

 We highlight axes of disparities on a global scale, continental scale and a more local scale. We intentionally do not look at one specific “homogenous” region on the continent in order to highlight unfairness that can affect the continent as a whole, as well as different levels of unfairness at the sub-regional and national levels. These axes of disparities are defined are defined with the African context in mind the logic behind it can be applied to different geographical regions. The remainder of the section discusses axes of disparities for fairness between Africa and non-African countries, and considerations for fairness attributes in the context of Africa. Individuals may have multiple axes of disparities. We define key terms in Appendix \ref{A1}.

\subsection{Axes of disparities between Africa and Western Countries}
\textbf{Colonial history:} All countries in Africa with the exception of Ethiopia and Liberia were colonized by Europeans. There is strong evidence that Africa's colonial history, and resulting structural challenges continue to create power imbalance between the colonized and colonial rulers and beneficiaries \citep{eichbaum21}. 
Colonial history has been put forward by several scholars as a social determinant of health \citep{turshen77, ramos22,czyzewski11}. This has implications for data collection. We propose colonial history as a fairness attribute that could cause historical, representation, measurement, learning, evaluation and deployment biases, and one that should be considered throughout the ML pipeline: problem formulation, data collection, model development and deployment. Learning bias is especially of note as objective measures for model performance should not be limited to accuracy, but should also seek to improve impact metrics. \\
 
\textbf{National income level (NIL):} Colonialism and its resulting structural issues link to national income level. The number of African countries by NIL are:  24 Low-income, 17 lower-middle income, 6 upper-middle income, and 1 high-income \citep{nada22}. 
Socio-economic disparities resulting from the disproportionately high number of LMICs on the continent imply limited funding from governments for research in ML for health, and limited availability of clinical data generating devices  and resources for ML development \citep{nabyonga21}. \\
This results in problem definitions that are more in line with external funding requirements, than with local needs. It also impacts access to data generation tools, which can lead to representation bias, as well as limited technical and personnel resources for ML development.\\ 

\textbf{Country of origin:}
Africa is divided into 5 distinct sub regions: North Africa, West Africa, Central Africa, East Africa and South Africa. Countries in these regions have different NIL levels, colonial histories, languages, cultures, ethnicity and racial subgroups. Country lines are colonial constructs, partitions resulting from the scramble for Africa, with the exception of Ethiopia and Liberia. This has implications for ethnicity, resources and development \citep{michalopoulos16}. Different countries may have varying methods of implementing health strategies. Hence machine learning applications that work in one country, may not necessarily transfer to the other. Aggregate bias which may result in a one-size fits all model should be avoided by understanding implications for fairness that may be country-specific. \\

\subsection{Globally applicable axes of disparities in the African context}
These disparities should be considered both in importing machine learning models from outside Africa, as well as developing machine learning models within the continent.
Disparities within the continent can result in different types of biases.\\

\textbf{Race:} Race in fairness typically refers to structural racism referenced as a social construct due to the history of slavery and subsequent racial disparities in countries like the United States, which does not apply directly to Africa. 
Majority of Africans (80\%) are racially black and are subjugated to global anti-black rhetorics \citep{pierre13}. North African Arabs are externally racially white,  as are Afrikaans in South Africa. There are also people of Asian origin, predominantly in the east and southern Africa.

However most Africans do not self-identify by race and are more likely to identify by nationality and ethnicity \citep{asante12, maghbouleh22}. Demographic health questionnaires do not typically ask for ``race", though persons of African descent may be at higher risk for certain diseases. If a model developed in the United States performs well across different races, it cannot be assumed that the model will also perform well across race, outside the United States, as there may be different underlying causes of disparities and could result in aggregation bias.

One may consider racial disparities that impact Africans on a global scale, or within different countries in Africa for example, between Black and White North Africans \citep{pierre13}, or between Black and White South Africans \citep{verwey12}. This has implications for socio-economic disparities in health access, representation bias and for race-based disparities in ML model performance.\\

\textbf{Ethnicity:} Africa has over 3000 ethnic groups and is the most genetically diverse continent. Unlike in HICs, most Africans associate with an ethnic group. Ethnicity, in addition to physical traits, language and culture, defines history, aspirations and world view \citep{deng97}. \\
 
Socio-economic disparities and approaches to accessing health may run along ethnic lines \citep{franck12}. This has implications that may skew ML applications towards dominant ethnic groups resulting in ethnic inequities. \\

\textbf{Religion:}
Africans practiced traditional religions pre-colonialism. With the introduction of Western religion, Christianity and Islam have grown to become dominant regions on the continent. Some countries are majority Christian with minority Muslim and traditionalists and vice versa. Religion can be associated with ethnicity and socio-economic status, leading to ``ethno-religious disparities" in health \citep{gyimah06, ha14}. Religion, can also impact access, perceptions and adoption of health practices \citep{white15, schmid08, obasohan14}. \\

\textbf{Language:}
There are over 2000 languages in Africa, and most Africans speak more than one \citep{heine2000}. Oral language in Africa typically has the format of a dominant colonial or ethnic language such as English, French, Arabic or Portuguese. There is usually a dominant native language, and several local dialects of less dominance. 
These dominant languages, particularly colonial languages, facilitate cross-ethnic interaction are typically both written and oral and are utilized in education and professional life \citep{obondo07}. 
There exist oral and written language barriers for people who speak less known dialects and for those with lower education and literacy levels, which can impact communication for health delivery \citep{levin06}.\\ 

\textbf{Skin tone:}
On a global scale, African skin tones, light or dark, are associated with blackness and instigate anti-blackness discrimination \citep{pierre13}.

Among black Africans, lighter skin tone may be perceived as enabling better social status and favors \citep{Nyoni21}. As a result 40\% of people on the continent, practice cosmetic skin bleaching \citep{asumah22}.

Bleaching in itself is a public health concern, increasing risk for severe health problems \citep{lewis12}. Within ML for health this has implications for practitioner attitudes, dermatology, such as skin disease detection, and measurement bias from devices that perform poorly on darker skin. \\

\textbf{Gender:}
Historically patriarchal African culture, reinforced by colonial legacies, has led to inequitable distribution of wealth between genders \citep{jaiyeola20}. Female household + child caring duties, and professional duties more recently further burdens and reduce access to basic health amenities for African women \citep{anaman20}. 
Transgender and gender non-conforming persons in Africa are stigmatized and discriminated against, with few health programs catering for transgender persons \citep{luvuno17}. This also has implications for health access, and treatment especially for high sexually transmitted diseases \citep{merwe20}.\\ 

\textbf{Sexual orientation:}
Thirty-two countries criminalize homosexuality and it is punishable by death in 3 countries \citep{lars22}. Homosexual persons are stigmatized and discriminated against which causes exclusion and marginalization within health systems \citep{sekoni22, ross21}. \\

\textbf{Literacy and Education level:}
Literacy and education levels impact access to care, health seeking behaviors and understanding of health information, especially when delivered digitally \citep{amoah18}. This has implications not only for one's own health, but also for the health of their children \citep{byaro21}.\\

\textbf{Age:}
Age is a global fairness attribute used both in machine learning model development and evaluations for health, due to age-specific incidence rates and co-morbidity. It is essential to demonstrate equitable ML model performance across age groups \citep{Mhasawade21}. \\

\textbf{Rural-urban divide:}
People living in rural areas may have disproportionate levels of lower socio-economic status, literacy, education and limited access to health facilities. This makes people in rural areas most vulnerable to unethical machine learning practices. However we must ensure just, beneficial ML applications that benefit rural regions, and improve health gaps.\\  

\textbf{Socio-economic status:} Individual socio-economic disparities underlie most of the above axes of disparities mentioned. Africa has the second highest wealth distribution gaps \citep{seery19}. This runs the risk of machine learning models in health perpetuating unfairness in deployment access or inaccuracies predominantly towards poorer persons.\\ 

\textbf{Disability:}
Ten to 20\% of African populations are affected by disabilities. Disabilities can exacerbate most of the attributes listed due to stigma and inadequate resources and policies at the country level \citep{adugna20, mckinney21}.\\

\textbf{Health-specific attributes:} 
Genetic and phenotype presentations, underlying pre-existing conditions, and later stages of disease presentation may impact model performance. For example, in breast cancer, people of African ancestry have higher prevalence of triple negative breast cancer.  In cervical cancer imaging, there may be varying cervix image presentations if a patient has co-existing conditions like HIV, or cervicitis. This directly impacts representation bias under disparate health-specific attributes.

\begin{figure*}[htbp]
\floatconts
  {fig:pipeline}
  {\caption{Biases along the machine learning and their associations with African-contextualized by axes of disparities.}}
  {\includegraphics[width=1\linewidth]{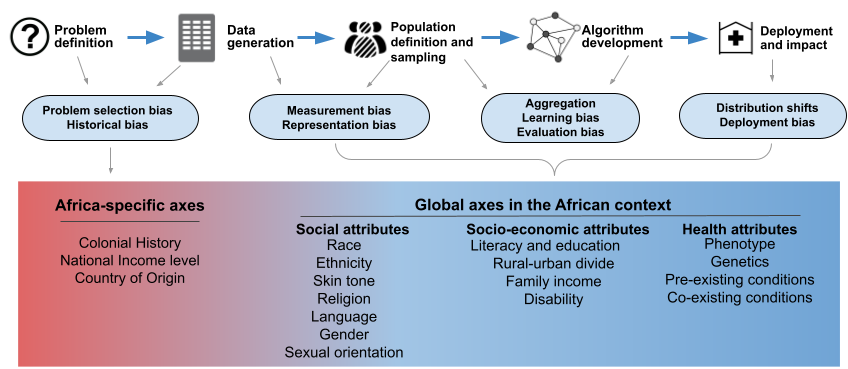}}
\end{figure*}

\section{Contextual barriers by health data modality}
\label{sec:math}

The field of ML for health relies on large open-source datasets. However most of these datasets are from countries outside of Africa. Though there are some open-source African health datasets, these are mostly limited to surveys due to limited use of electronic health records. We discuss how these can be impacted by fairness attributes across different health modalities. Definitions of the types of bias can be found in Appendix \ref{A1}.\\

\textbf{Medical imaging:} Machine learning has shown the most promise for use in medical imaging tasks such as dermatology, pathology, mammography, ultrasound, CT, MRIs and X-rays \citep{castiglioni21}. However imaging devices are limited in hospitals in Africa, due to their high cost, and maintenance. Even when imaging devices are available, they may be of varying quality and interpretation, which leads to poor generalization of ML models from HICs. If available, they may be at a higher costs and in urban areas, which leads to disparities by socio-economic status, and the rural-urban divide. Additionally medical images may not be digitally stored or connected to other patient health records. This can lead to limited data for retraining models and can generate representation, and evaluation biases.  \\

\textbf{Survey data:}
Demographic and health survey data present some of the largest, readily available longitudinal sources of health data from the continent. These surveys are used to provide data on disease trends over time by different regions, and demographics. While they allow data representation from African of different backgrounds, when using them for ML model development, care must be taken to ensure that models do not rely on proxy or sensitive attributes that could lead to unfairness towards certain demographic groups. This has implications for unfairness across several of the attributes listed including gender, sexual orientation, ethnicity, disability and  socio-economic status. \\

\textbf{Unstructured written health notes:} Machine learning for unstructured health notes utilizes large computer vision and natural language processing models  to extract information such as symptoms and action items, and provide disease classifications. These models are usually pre-trained on large amounts of text data from overserved languages. They already exhibit biases by culture, gender, and race. Using them without proper evaluation and fine-tuning may propagate these identified biases in addition to language-based biases. In addition  to the modelling challenges, there exist limitations to accessing written health notes in health facilities due to limited EHR availability. \\

\textbf{Medical speech:} Automatic speech recognition systems are used in various heath facilities by healthcare professionals to dictate notes without having to take time away from patient care. Accent, style of speaking (for instance pidgin) and literacy may impact speech recognition algorithms designed to be used in Africa. This can impact language disparities, and indirectly ethnicity, country of origin, literacy and education. \\

\textbf{Optical sensor devices:}
Optical sensor-based such as pulse oximeters and fitness trackers have been shown to have lower performance on darker skinned persons \citep{shi22}. Machine learning models developed for these devices may perpetuate measurement bias. This has direct implications for skin-tone disparities, which may be indirectly linked to country of origin, and ethnicity. \\

\textbf{Facial recognition:} Facial recognition algorithms have shown use in health to diagnose medical conditions such as autism \citep{liu16}. However these algorithms have also been shown to have high performance bias by skin tone and gender \citep{buolamwini18}. There also exists differences cultural differences in facial expression that must be taken into account \citep{dailey10}. \\

\textbf{Omics:}
 Using ML to discover biomarkers from multi-omic data is a fast growing field, with potential for precision diagnosis, prognosis and prediction \citep{reel21}. Africa has large genetic diversity, but available omics data is underrepresented, making up only 1\% of omics databases \citep{hamdi21}. While the amount of data is increasing, facilitated by consortiums such as Human Heredity and Health in Africa (H3Africa) consortium and the H3Africa Bioinformatics Network (H3ABioNet), there remain limitations in labelled data, ethnic diversity representation, and integration with a patient's health history and profile \citep{hamdi21}. \\

\textbf{Lab values:}
Depending on hospital facility resources, patients referred for labs can obtain them within the hospital, though they are typically referred to external laboratories for testing. Access to laboratory testing services may be especially limited for people with lower financial means, and those in rural areas who may have to travel to urban-based lab institutions \citep{petti06}. Results from these tests are provided to the patient who then provides it to the referring clinician. Given limited EHRs, these results are not always integrated with the patients comprehensive health profile leading to disaggregated health information. In addition to data access limitations, limited infrastructure and personnel (eg. pathologists) for lab interpretation impact diagnostic efficiency and accuracy \citep{petti06}. These present unique challenges for ML purposes which are not observed in HICs. \\

These modality-based contextual barriers have implications for upstream ML tasks in classification, medical image segmentation, automated speech recognition, natural language processing and health recommender systems which may lack Africa- specific datasets and contexts. \\

\begin{figure}[htbp]
\floatconts
  {fig:modality}
  {\caption{African-contextualized barriers to ML for health by health modality. *=applies to all}}
  {\includegraphics[width=1\linewidth]{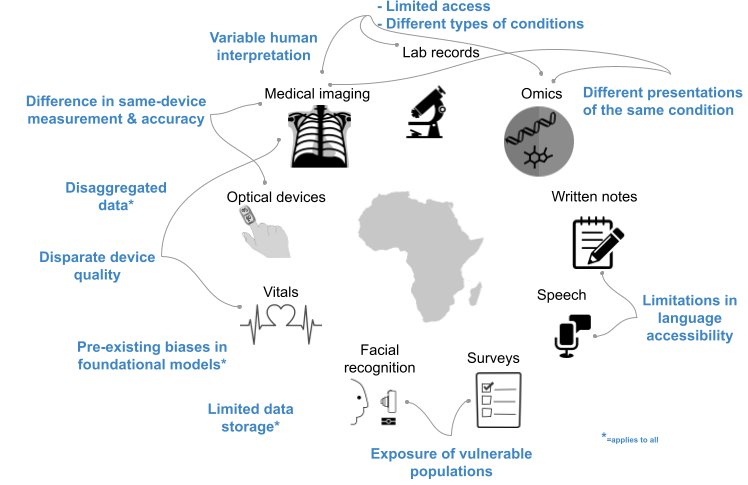}}
\end{figure}

\section{Implications for machine learning and research opportunities}
\label{sec:floats}

Several mitigation strategies have been proposed to address bias in machine learning models. Most methodological advances to date have focused on (1) defining a fairness criterion to be satisfied by a model and (2) devising procedures across the model development pipeline to guarantee such criterion. Given the inherent dependencies of these steps to the problem at hand, it is important to understand what particular challenges may arise when building and deploying models within the African context \citep{BEHARA2022e01360, Waljee1259}.\\

\textbf{Contextualization of the fairness criteria:} The initial formalization of popular algorithmic fairness metrics \citep{barocas-hardt-narayanan}, such as demographic parity and equality of opportunity, was intrinsically tied to anti-discrimination laws in the US \citep{barocas2016} around disparate impact and disparate treatment. This motivated the development of methodologies to ensure that automated systems did not have discriminatory effects based on contextually-relevant protected attributes such as race and gender \citep{Hutchinson2018}. As highlighted in previous sections, the diversity and axes of disparities between protected attributes in HICs and LMICs lead to a fundamental need to study which fairness definitions are pertinent within the local historical context. By virtue of the field still being in its infancy on the African continent, there is a unique opportunity to engage different stakeholders (researchers, policymakers, governance, etc) to define relevant fairness desiderata in accordance with local laws and beliefs. An interdisciplinary approach at this stage is particularly important to avoid potential issues around competing definitions of fairness \citep{verma2018, Kleinberg16, Chouldechova17} and purely mathematical formulations that lead to unintended performance degradation for all groups \citep{levellingdown2023}.\\

\textbf{Practical fairness considerations:} Further studies in the algorithmic fairness field have highlighted the different ways in which bias can manifest in each step of the ML development pipeline namely during (1) problem selection, (2) data collection and processing, (3) algorithm development and evaluation and (4) post deployment considerations \citep{chenIY21}. 
A plethora of debiasing approaches have been proposed in the literature to tackle potential biased during steps (3) through (5) \citep{Mehrabi2021, barocas-hardt-narayanan}. 
In addition to better understanding how these mitigation strategies should be adapted to account for varying fairness attributes specific to LMICs, a particular hindrance to the application of machine learning techniques to the health domain in Africa resides in the problem formulation, collection and utilization of data \citep{Okolo2023}. 
Below, we highlight examples of these challenges and important fairness considerations when using machine learning tools in Africa. 

\begin{itemize}
    \item {\it problem selection:} As noted in \citep{chenIY21}, biases stemming from unaligned incentives by external organizations can severely affect, for example, which diseases are studied. This has a trickling effect in terms of which data is collected regardless of relevance to the local population. It is particularly important at this step to consider the potential biases of stakeholders and ensure affected communities are included in the decision making step. 
    \item {\it scarcity of digitalized health records:} patients health records remain predominantly paper-based due to a lack of infrastructure and computer resources to keep digital copies. This phenomenon is exacerbated in rural areas leading to further inequalities. Awareness of the crucial role of having electronic records in order to utilize machine learning tools ushers in new proposals to digitalize existing handwritten data. Challenging fairness problems in this step include how to develop automated tools that can reliably process underserved African languages as well as  how to ensure that such efforts are equally distributed across regions.
    \item {\it prevalence of survey data:} most application of machine learning to health remain tied to data collected in clinical settings \citep{Mhasawade21}. In the African context, the prevalence of self-reported survey data introduces a number of biases including unbalanced representation of groups based on who answers the survey and potentially incorrect self-reported information about a responder's attributes. Application of machine learning methods to such data modality and how to account for measurement biases remain an underexplored area of research.\\
\end{itemize}

\textbf{Caution around using pretrained models:} Given the scarcity of readily available large training datasets from Africa, a common approach consist of finetuning large pretrained models to downstream tasks. 
This approach is particularly tempting as it allows to harness the power of large models whose training requires considerable resources and infrastructure within more resource-constrained environments. 
However, there are several pitfalls in this approach that could lead to unintended biases in the final model. 

A prevalent concern lies within the inability to adapt fairness properties to distribution shifts which are likely to occur when deploying models off-the-shelf or finetuning to a target dataset. However, even if bias mitigation techniques are used at training time, there are no guarantees that the fairness property will hold after deployment or after the finetuning process. Below we highlight some examples of real world distribution shifts within the African context.

\begin{itemize}
    \item {\it Demographic shifts} occur when the distribution of the fairness attribute changes between training and test time. This is likely to happen when deploying models trained on Western centric data to African data given the representation of race groups for example is likely to vary between the two datasets. 
    \item {\it Covariate shifts} happen when the distribution of input features changes between train and test data. For instance ML-enabled dermatology diagnosis, may have been trained on lighter skin and may not work on presentations of the same disease on dark skin. Another example is the stage at which disease is presented. Given limited preventative screening, diseases may be diagnosed at later stages than in HICs, affecting data distribution for model classification. 
    \item {\it Label shifts} relate to changes in the distribution of outcomes between the training data and test data.  This type of shifts can be seen in differences in health conditions, i.e. diseases that may be specific to Africa and that are undersampled in training datasets from Western countries.
\end{itemize}
 Ensuring the generalizability and transferability of fairness properties across domains and under distribution shifts is a new and active area of research  \citep{schrouff2022diagnosing, giguere2022fairness, singh21, baldini-etal-2022-fairness, sadeghi20}. Use-cases specific to the African context can serve as motivating examples and drive impactful advances in this area.   \\

While we discuss ML based mitigation strategies in this section, addressing machine learning biases starts from structural, and systemic changes in healthcare in Africa. Access to basic health care for all persons; increased government and private investment in health access, health-based entrepreneurship and health-related research; and sustainable, and context-specific development and deployment of health tools and technologies are essential for reducing disparities in health.  

\section{Conclusions}
 There are unique opportunities for machine learning to make positive impact in Africa and advance global health equity. However there should be proactive steps taken to prevent harms, reduce biases, and ensure fairness. To develop fair machine learning models, one needs first to understand what the fairness attributes of a given context are and where to apply them. We provide this work as a starting point towards building the foundation for fairness in machine learning in global health with a focus on Africa.  Future work will involve qualitative and quantitative analysis of machine learning models, how they respond to these attributes, and mitigation proposals. 

\bibliography{jmlr-sample}

\appendix

\label{A1}

\section{Definitions}
We define terms that are used in this paper. While there is not a standardized definition of fairness, we mainly follow the taxonomy described in \citep{Mehrabi2021}. We also provide other terms of general informational value for further understanding of fairness.\\

\textbf{Discrimination:}
`Discrimination can be considered as a source for unfairness that is due to human prejudice and stereotyping based on the sensitive
attributes, which may happen intentionally or unintentionally'.\\

\textbf{Bias:}
`Bias can be considered as a source for unfairness that is due to the data collection, sampling, and measurement.'\\

\textbf{Types of Bias} \citep{suresh21}

\textit{Historical Bias:}
`Historical bias arises even if data is perfectly measured and sampled, if the world as it is or was leads to a model that produces harmful outcomes. Such a system, even if it reflects the world accurately, can still inflict harm on a population. Considerations of historical bias often involve evaluating the representational harm (such as reinforcing a stereotype) to a particular group'.\\

\textit{Representation Bias:}
`Representation bias occurs when the development sample underrepresents some part of the population, and subsequently fails to generalize well for a subset of the use population'. \\

\textit{Measurement Bias:}
`Measurement bias occurs when choosing, collecting, or computing features and labels to use in a prediction problem. Typically, a feature or label is a proxy (a concrete measurement) chosen to approximate some construct (an idea or concept) that is not directly encoded or observable. Proxies become problematic when proxy's are an oversimplification, are measured differently across groups, or accuracy differs across groups'. \\

\textit{Aggregation Bias:}
`Aggregation bias arises when a one-size-fits-all model is used for data in which there are underlying groups or types of examples that should be considered differently.A particular data set might represent people or groups with different backgrounds, cultures, or norms, and a given variable can mean something quite different across them. Aggregation bias can lead to a model that is not optimal for any group, or a model that is fit to the dominant population'. \\

\textit{Learning Bias:}
`Learning bias arises when modeling choices amplify performance disparities across different examples in the data. Issues can arise when prioritizing one objective (e.g., overall accuracy) damages another (e.g., disparate impact)'. \\

\textit{Evaluation Bias:}
`Evaluation bias occurs when the benchmark data used for a particular task does not represent the use '.\\

\textit{Deployment Bias:}
`Deployment bias arises when there is a mismatch between the problem a model is intended to solve and the way in which it is actually used'. \\

\textbf{Fairness}: 
`In the context of decision-making, fairness is the absence of any prejudice or favoritism toward an individual or group based on their inherent or acquired characteristics'. \\

\textbf{Types of Fairness}

\textit{Individual Fairness} \citep{dwork2012}:
`Individual fairness is captured by the principle that any two individuals who are similar with respect to a particular task should be classified similarly'.\\

\textit{Counterfactual Fairness} \citep{kusner2017}
`The counterfactual fairness definition is based on the “intuition that a decision is fair towards an individual if it is the same in both the actual world and a counterfactual world where the individual belonged to a different demographic group.'\\

\textit{Group Fairness:} 

Broadly, group fairness notions aim to `treat different groups equally'. Below are two main statistical group fairness definitions that exist in the literature. 
 
 \begin{itemize}
     \item \textit{Equal Opportunity} \citep{hardt2016}:
 `This fairness notion requires that the probability of a person in a positive class being assigned to a positive outcome to be equal for both protected and unprotected (female and male) group members. In other words, the equal opportunity definition states that the protected and unprotected groups should have equal true positive rates'.
 \item \textit{Demographic Parity} \citep{dwork2012}:
` requires that the overall proportion of individuals in a protected group predicted as positive (or negative) to be the same as that of the overall population'. \\
 \end{itemize}

\textit{Fairness in Relational Domains:}
`A notion of fairness that is able to capture the relational structure in a domain—not only by taking attributes of individuals into consideration but by taking into account the social, organizational, and other connections between individuals'. \\

\textit{Subgroup Fairness:}
Subgroup fairness intends to obtain the best properties of the group and individual notions of fairness. It picks a group fairness constraint like equalizing false positive and asks whether this constraint holds over a large collection of subgroups. \\

\textbf{Axes of Disparities}\\

\textit{Africa-specifc axes of Disparities:}
Disparities that primarily impact Africa. \\

\textit{Global axes of disparities in the context of Africa:}
Disparities that have global implications but contextualized for Africa.

\end{document}